\documentclass{article}
\usepackage{amssymb}
\usepackage{marvosym}
\usepackage{textcomp}

\PassOptionsToPackage{numbers, compress}{natbib}
 \usepackage[preprint]{neurips_2026}

\usepackage[utf8]{inputenc} 
\usepackage[T1]{fontenc}    
\usepackage{url}            
\usepackage{booktabs}       
\usepackage{amsfonts}       
\usepackage{nicefrac}       
\usepackage{microtype}      
\usepackage{xcolor}         
\usepackage{wrapfig}
\usepackage{colortbl}       
\usepackage{makecell}       

\definecolor{secHead}{HTML}{2C3E63}   
\definecolor{rowOurs}{HTML}{F4ECD2}   
\definecolor{ruleSoft}{gray}{0.85}    

\usepackage{graphicx}
\usepackage{amsmath, amssymb, amsthm}
\usepackage{caption}
\usepackage{subcaption}
\usepackage[colorlinks=true, allcolors=blue!60!black]{hyperref}
\usepackage[capitalize]{cleveref}
\usepackage{enumitem}
\usepackage{algorithm}
\usepackage{fancyvrb}

\newcommand{\sff}{\textsc{SF}}
\newcommand{\jit}{JiT}
\newcommand{\fid}{FID}
\newcommand{\is}{IS}
\DeclareMathOperator{\DCT}{DCT}
\DeclareMathOperator{\IDCT}{IDCT}

\title{Show the Signal, Hide the Noise:\\
Spectral Forcing for Pixel-Space Diffusion}

\author{Weichen Fan$^{1}$ \hspace{.2cm}  Haiwen Diao$^{1}$ \hspace{.2cm} Penghao wu$^{1}$  \hspace{.2cm}  Ziwei Liu$^{1,}$\textsuperscript{\Letter} \\
\textsuperscript{1}S-Lab, Nanyang Technological University \\
{\tt\small {\{weichen002,penghao001\}@e.ntu.edu.sg, \ \{haiwen.diao,ziwei.liu\}@ntu.edu.sg}}\\[1ex]
\parbox{\textwidth}{
\centering
\begin{tabular}{ll}
\raisebox{-0.15em}{\includegraphics[height=1.05em]{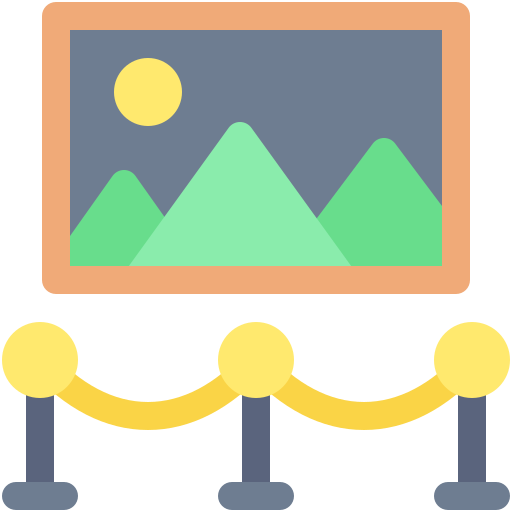}} \textbf{Github:} & \url{https://github.com/WeichenFan/Spectral_Forcing}. \\
\raisebox{-0.15em}{\includegraphics[height=1.05em]{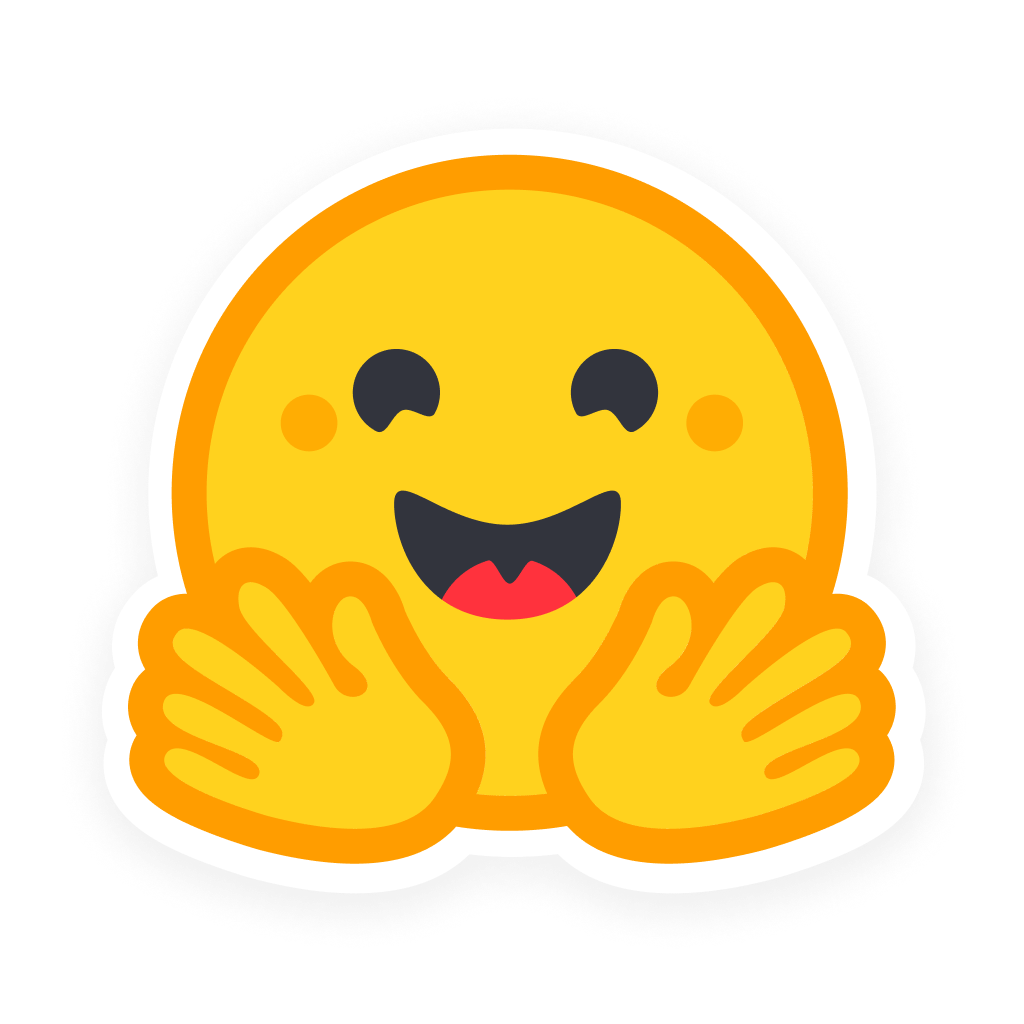}} \textbf{Hugging Face:} & \url{https://huggingface.co/weepiess2383/Spectral_Forcing}. \\
\end{tabular}
}
}
\begin{document}

\maketitle

\vspace{-20pt}
\begin{abstract}
Pixel-space diffusion models are trained on full-bandwidth noisy images, yet the useful signal available to the denoiser is strongly frequency dependent. Under rectified-flow diffusion and natural-image power-law spectra, the per-band data-to-noise contour $k^{\ast}(t)=(1-t)^{-2/\alpha}$ separates a signal-bearing low-frequency region from a noise-dominated high-frequency region at each time $t$. We show that this implicit coarse-to-fine structure is not merely descriptive: it induces a capacity-allocation problem. A standard pixel-space denoiser must discover the moving bandwidth boundary internally and can spend computation on frequency-time regions where the optimal prediction collapses to deterministic baselines rather than data-distribution modeling. To make this boundary explicit, we introduce \emph{Spectral Forcing}, a parameter-free, time-conditional 2D-DCT low-pass operator applied to the noisy input before the patch embedder. Its cutoff expands monotonically with the diffusion time and becomes the identity at the data endpoint. Through controlled synthetic experiments, we identify the regime in which the operator is beneficial: coarse patch tokenization and data whose high-frequency content is predominantly noise rather than essential signal. On ImageNet-256 with JiT-700M/32, Spectral Forcing consistently improves both FID and Inception Score across different training epochs, demonstrating robust gains throughout training; at finer tokenization, the spectral forcing is still competitive. We further insert the unchanged operator into SenseNova-U1, a unified text-to-image model, where it improves DPG-Bench and GenEval, showing that the input-side spectral prior transfers beyond class-conditional generation. These results suggest a route to capacity-efficient pixel-space diffusion by showing the signal and hiding the noise.
\end{abstract}
\section{Introduction}
\label{sec:intro}

\begin{figure}[t]
\centering
\includegraphics[width=\textwidth]{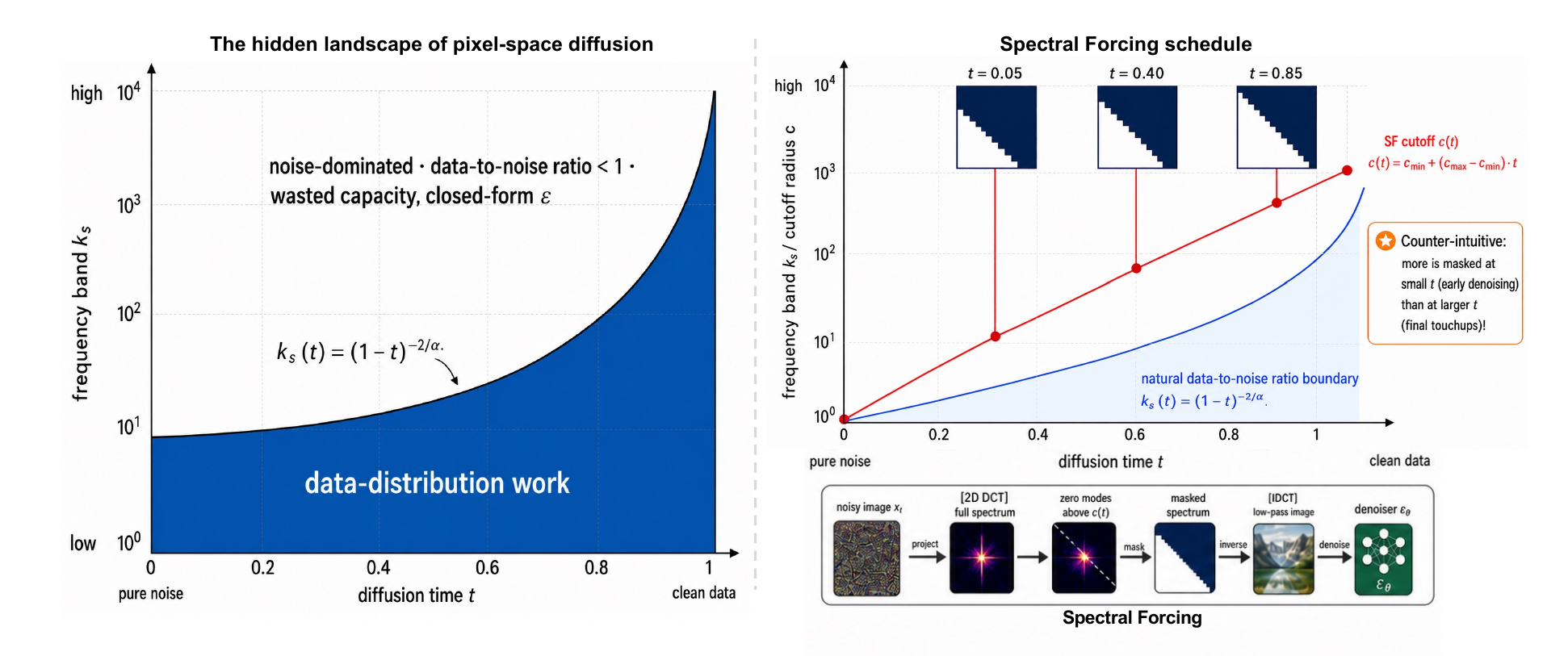}
\caption{
\textbf{Spectral Forcing for pixel-space diffusion.}
\textit{Left:} the per-band data-to-noise contour $k^{\ast}(t){=}(1{-}t)^{-2/\alpha}$ separates a signal-bearing region (data-distribution work) from a noise-dominated region where an unforced denoiser collapses to a closed-form map (wasted capacity).
\textit{Right:} \sff{} imposes the boundary explicitly with a parameter-free, time-conditional 2D-DCT low-pass at cutoff $c(t)$, applied before the patch embedder; $c(t)$ grows monotonically with $t$ and is the identity at $t{=}1$. Bottom strip: one operator step --- noisy input $\to$ 2D-DCT $\to$ mask above $c(t)$ $\to$ IDCT $\to$ denoiser $\varepsilon_\theta$. The diffusion objective, architecture, and sampler are unchanged.
}
\label{fig:teaser}
\end{figure}

Diffusion and flow-based models are the state of the art for generating high-quality images. Until recently, the dominant recipe was to operate in a compressed latent space produced by a separately trained autoencoder, with the diffusion model itself learning to denoise latents rather than pixels. This separation has been justified on practical grounds (latents are smaller and faster to denoise), but it adds an external dependency to the generative recipe and obscures the spectral structure of the underlying images. Recent work shows that pixel-space diffusion can be competitive when the architecture is properly designed, in particular through coarse patch tokenization and large transformer backbones~\citep{li2025jit, lin2024dctdiff, wang2024pixnerd}.

A common observation across these recipes is that diffusion training is implicitly coarse-to-fine: at each timestep the noise level determines a frequency band above which the data signal is buried in noise, and the network must learn to reconstruct lower-frequency content first and higher-frequency content last~\citep{dieleman2024spectral}. This implicit hierarchy has been documented through frequency-content analyses but has rarely been exploited as an explicit architectural prior. We argue that the hierarchy is not merely descriptive but induces a capacity-allocation problem: a standard pixel-space denoiser, faced with the full bandwidth of the noisy input at every timestep, must discover the moving bandwidth boundary internally and can spend computation on frequency-time regions where the optimal prediction collapses to deterministic baselines rather than data-distribution modeling.

The reason this hierarchy emerges is straightforward. Under rectified-flow diffusion, the network at time $t$ observes $z_t = t\,x + (1-t)\,\varepsilon$ with $\varepsilon \sim \mathcal{N}(0, I)$. For natural-image-like data with power spectrum $P(k) \propto k^{-\alpha}$, the per-band data-to-noise ratio is $k^{-\alpha}/(1-t)^2$ (the data spectrum compared to the additive-noise variance floor; see \cref{sec:method:prelim} for the relation to the standard $z_t$-SNR), and the contour $\mathrm{DNR}(k, t) = 1$ defines a moving cutoff $k_*(t) = (1-t)^{-2/\alpha}$ that separates a signal-bearing region from a noise-dominated region (\cref{fig:teaser}, left). The standard network has no architectural awareness of this cutoff: it must identify it implicitly from the noise schedule, and allocate capacity between learning data-distribution structure where signal exists and reproducing deterministic baselines where it does not. We confirm this allocation directly with a per-band MSE diagnostic at convergence on synthetic data (\cref{sec:method:empirical}): the network does meaningful data-distribution work only in a wedge of $(t, k)$ space, and converges to deterministic baselines elsewhere.

We ask: \emph{can making the bandwidth boundary explicit at the input free model capacity for the harder part of the task?} We introduce \textbf{Spectral Forcing} (\sff{}), a parameter-free time-conditional 2D-DCT low-pass mask applied to the network input before the patch embedder (\cref{fig:teaser}, right). The mask's cutoff radius $c(t)$ grows monotonically with the diffusion timestep along a fixed-by-design schedule that is imposed at the input rather than estimated from the data spectrum, restricting the network's view of $z_t$ to the bands where signal can dominate, and saturating to the identity at the data endpoint so the trajectory still integrates full-bandwidth velocity. The operator introduces no learnable parameters, costs about half a percent of total compute at $256^2$, and composes with any pixel-space rectified-flow recipe without modifying the forward process, the loss, the EMA, the sampler, or classifier-free guidance.

\sff{} is regime-dependent. We show, both in toys and on ImageNet, that the operator delivers its largest gains in a specific conjunction of conditions: (i) the model's patch tokenization is coarse enough that the patchify already aggressively bandlimits the representation, and (ii) the data's high-frequency content is predominantly noise rather than essential signal. When these conditions hold, the operator delivers consistent improvements throughout training; when they do not, it remains competitive with the unforced baseline. We make this regime explicit through a controlled toy experiment so that practitioners know where to expect headline gains versus where the operator's main role is as a non-harmful regularizer. The coarse-tokenization regime is also the operating point of recent native vision-language models~\citep{diao2025neo} that bypass an external visual encoder and process raw image patches directly, where token count must be kept small for tractable joint sequence modelling; making capacity-efficient pixel-space diffusion practical in this regime is therefore a downstream-relevant target.

On ImageNet-256 the empirical headline is sharp. At the JiT-700M/32 configuration (64 transformer tokens, the largest configuration in~\citet{li2025jit}), \sff{} reduces FID from $24.19$ to $20.68$ ($+14\%$) and improves Inception Score from $83.28$ to $93.96$ ($+13\%$) in an apples-to-apples 60-epoch comparison against a same-recipe baseline. The improvement is robust across training budgets: \sff{} consistently improves both FID and Inception Score at every epoch checkpoint we evaluate, demonstrating that the gain is not a transient data-efficiency artifact. At finer tokenization (256 tokens, JiT-130M/16), \sff{} remains competitive with the baseline at the matched 60-epoch budget, delimiting the regime where it produces additional headline gains versus where it serves as a non-harmful frequency prior.

Our work makes the following contributions:
\begin{itemize}[leftmargin=*, itemsep=2pt, topsep=2pt]
\item We formalize a per-band data-to-noise analysis as a closed-form bandwidth-coherence framework that predicts the operator's optimal cutoff schedule, the analytical bandwidth front $c(t) \propto (1-t)^{-2/\alpha}$, and identifies its regime of applicability.
\item We design a controlled 1D-to-2D toy experiment that derives the operator from a per-band MSE diagnostic (showing that an unforced network has converged to a deterministic baseline outside a wedge of $(t, k)$ space), characterizes its dependence on patch size and data spectrum, and exposes when it helps or hurts. We verify the same wedge structure directly on real ImageNet checkpoints.
\item We validate on ImageNet-256: a $+14.5\%$ FID and $+13\%$ Inception Score gain at \jit-700M/32 in an apples-to-apples $60$-epoch comparison, holding $+8.0\%$ at $120$ epochs where \sff{} already matches a published $\sim$$145$-epoch reference. Our method surpasses constant low-pass, spatial Gaussian blur, Focal Frequency Loss~\citep{jiang2021focal}, blurring diffusion~\citep{hoogeboom2023blurring}, and DCTDiff~\citep{lin2024dctdiff} at the same operating point.
\end{itemize}

Together these results suggest a simple route to more capacity-efficient pixel-space diffusion by showing the denoiser the signal and hiding the noise.

\section{Related Work}
\label{sec:related}

\paragraph{Diffusion and pixel-space generation.}
Diffusion and flow-matching dominate high-quality image generation~\citep{sohl2015deep,ho2020denoising,nichol2021improved,song2020score,song2021scorebased,karras2022edm,lipman2022flow,liu2023rectified,albergo2022stochastic}, typically with transformer backbones~\citep{dosovitskiy2020vit,peebles2023dit,ma2024sit} replacing U-Nets~\citep{ronneberger2015unet}. Latent diffusion compresses images via separately trained autoencoders~\citep{rombach2022ldm,podell2023sdxl,esser2024scaling,bfl2024flux}, with recent work substituting representation-rich tokenizers for faster convergence~\citep{yu2024repa,yao2024vavae,zheng2025rae,shi2025svg,yue2025uniflow,fan2026uae}. Pixel-space diffusion is the alternative~\citep{dhariwal2021adm,saharia2022imagen,nichol2022glide,ho2022cascaded,hoogeboom2024simple,simo2024pixelflow,wang2024pixnerd,li2024mar,nguyen2024pixeltransformer}; we build on JiT~\citep{li2025jit}, where large-patch transformers match latent baselines without auxiliary losses. The coarse-tokenization regime is also the operating point of native VLMs~\citep{diao2025neo} that process raw image patches without a separate encoder. \sff{} is a parameter-free input-side adapter that composes with any of these and leaves forward process, loss, and sampler unchanged.

\paragraph{Spectral and frequency-domain methods.}
Some prior work makes the forward process itself spectral, replacing Gaussian noise with progressive blurring or wavelet shrinkage~\citep{rissanen2022ihdm,hoogeboom2023blurring}; analyses without modifying the forward process formalize standard diffusion's implicit coarse-to-fine character~\citep{dieleman2024spectral} and link it to the spectral bias of neural networks~\citep{rahaman2019spectral,tancik2020fourier,sitzmann2020siren} and natural-image power-law statistics~\citep{torralba2003statistics,burton1987color,ruderman1994statistics}. Other work generates directly in a frequency representation~\citep{lin2024dctdiff} or coarse-to-fine token order~\citep{huang2025nfig,wang2025nvg,tian2024var,denton2015laplacian,fan2023frido,yellapragada2025zoomldm,skorokhodov2024hierarchical}, sometimes with frequency-aware objectives~\citep{jiang2021focal,karras2021aliasfree}. Latent Forcing~\citep{baade2026latentforcing} cascades a frozen semantic encoder with a pixel-level diffusion head. \sff{} differs structurally: the forward process is unchanged rectified-flow, the architecture is unchanged, and the operator is a parameter-free time-conditional mask on the pixel input whose schedule is derived from the per-band data-to-noise contour of the unmodified forward process.

\section{Methodology}
\label{sec:method}

\subsection{Preliminaries}
\label{sec:method:prelim}

\paragraph{Rectified-flow.}
For data $x \sim q(x \mid y)$ and noise $\varepsilon \sim \mathcal{N}(0, I)$, rectified-flow~\citep{liu2023rectified, lipman2022flow} linearly interpolates between source and data:
\begin{equation}
z_t \;=\; t\,x + (1 - t)\,\varepsilon, \qquad t \in [0, 1],
\label{eq:forward}
\end{equation}
so $z_0 = \varepsilon$ is noise and $z_1 = x$ is the data. The per-sample velocity target is
\begin{equation}
v_{\text{target}} \;=\; \frac{x - z_t}{1 - t} \;=\; x - \varepsilon,
\label{eq:vtarget}
\end{equation}
and a flow-matching model $v_\theta(z_t, t, y)$ is trained against $v_{\text{target}}$ under squared error,
\begin{equation}
\mathcal{L}(\theta) \;=\; \mathbb{E}_{x, \varepsilon, t}\big[\| v_\theta - v_{\text{target}}\|^2\big],
\label{eq:loss}
\end{equation}
with $t$ logit-normal at training and EMA weights at inference. Sampling integrates $v_\theta$ from $t{=}0$ to $1$ with a Heun integrator and classifier-free guidance.

\paragraph{Per-band data-to-noise ratio.}
We approximate the radial 2D-DCT spectrum of natural images by
$P(k)\propto k^{-\alpha}$, with $\alpha\approx2.82$ on ImageNet-256
(\cref{app:alpha}). Since \cref{eq:forward} adds per-band noise variance
$(1-t)^2$, we define the per-band \emph{data-to-noise ratio} (DNR) as
\begin{equation}
\mathrm{DNR}(k,t)
=
\frac{P(k)}{(1-t)^2}
=
\frac{k^{-\alpha}}{(1-t)^2}.
\label{eq:dnr}
\end{equation}
Its unit level set gives a closed-form bandwidth front,
$k_*(t)=(1-t)^{-2/\alpha}$, above which noise dominates the data spectrum.
DNR references the clean-data power $P(k)$ rather than its attenuated image
$t^2 P(k)$ in $z_t$, and relates to the conventional $z_t$-SNR by
$\log\mathrm{SNR}=\log\mathrm{DNR}+2\log t$. The $2\log t$ term is constant in
$k$: it scales every band identically and leaves the spectral slope $-\alpha$
unchanged, so it carries no frequency-discriminative information and shifts only
the absolute level at which the ratio crosses unity, which we fold into the
cutoff $c(t)$.


\begin{figure}[!t]
\centering
\includegraphics[width=\textwidth]{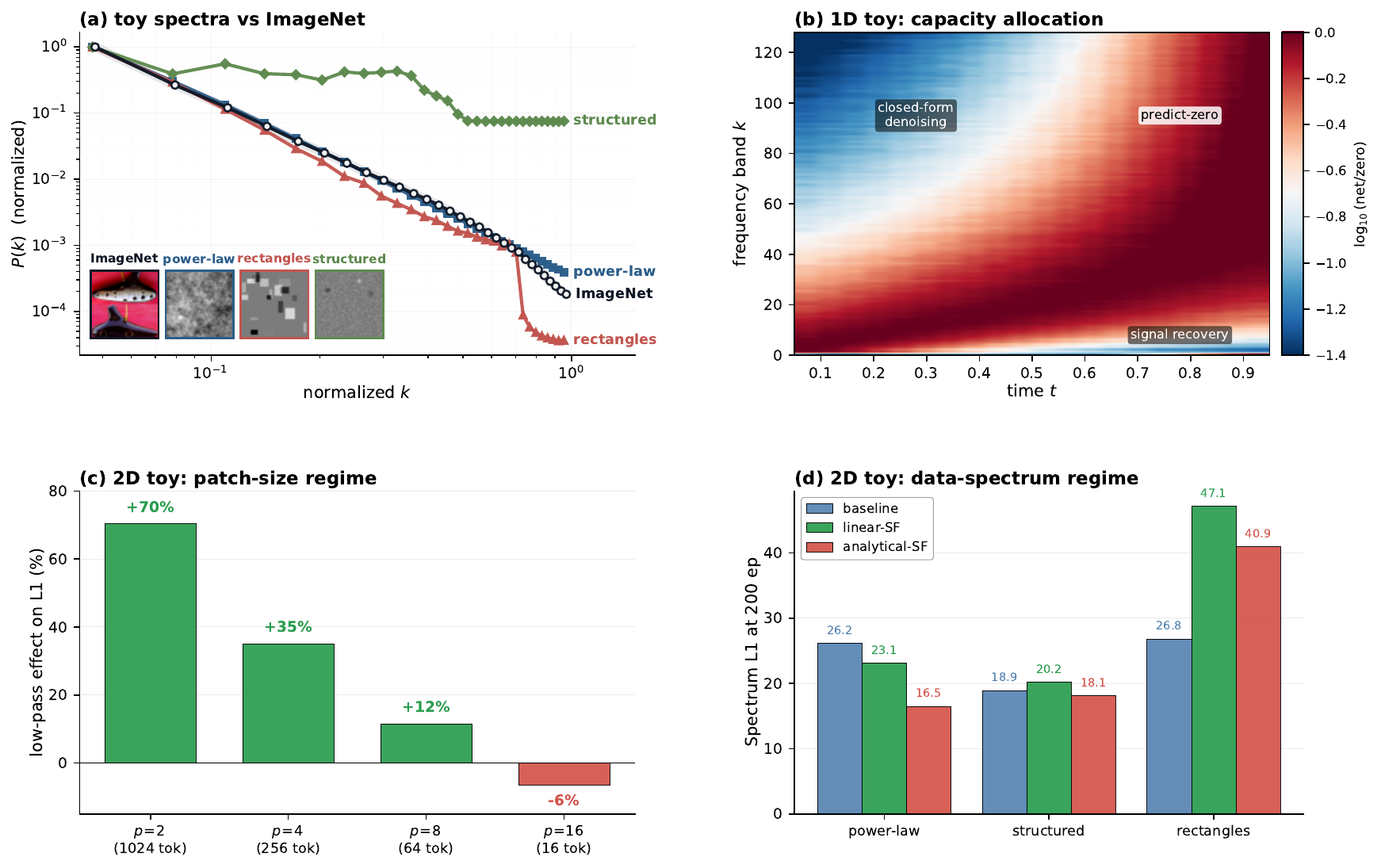}
\caption{
\textbf{Three empirical motivations for Spectral Forcing.}
\textbf{(a)} Radial 2D-DCT power spectra of the three toy distributions, overlaid on ImageNet-256 (insets: samples).
\textbf{(b)} Converged 1D toy denoiser: per-band $\log_{10}(\mathrm{MSE}_{\mathrm{net}} / \mathrm{MSE}_{\mathrm{zero}})$ on the $(t, k)$ plane reveals three regions: \emph{signal recovery} (low-$k$ wedge, the only region of true data-distribution work), \emph{closed-form denoising} (low $t$, high $k$), \emph{predict-zero} (high $t$, high $k$).
\textbf{(c)} 2D toy DiT: input-side time-conditional low-pass vs.\ patch size ($h{=}64$, $\alpha{=}2$). Helps at coarse $p$; starves at very fine $p$.
\textbf{(d)} Same operator across data spectra at $p{=}8$. Helps on power-law (analytical $\gg$ linear), neutral on structured, hurts on rectangles where high-frequency content is essential signal.}
\label{fig:method_empirical}
\end{figure}

\subsection{Empirical Study}
\label{sec:method:empirical}

The bandwidth front $k_*(t)$ partitions the $(k, t)$ plane into signal-bearing (below) and noise-dominated (above) regions; as $t \to 1$ the front sweeps outward, exposing more bands. The empirical question is whether a standard pixel-space denoiser uses this structure, and where the allocation becomes wasteful. We answer with three controlled experiments on small models (\cref{fig:method_empirical}), before any version of Spectral Forcing is introduced: a 1D rectified-flow Transformer (${\sim}178$k params) on synthetic 1D power-law signals, and a 2D DiT (${\sim}3$M params) on $h{\times}h$ synthetic images at $h{=}64$, trained under the recipe of \cref{eq:loss}.

\noindent\textbf{Result 1: the network does data-distribution work only in a wedge.}
At convergence we measure the 1D model's per-band velocity-prediction MSE relative to the trivial zero-predictor baseline $\|v_{\mathrm{target}}\|^2$ on a dense $(k, t)$ grid (\cref{fig:method_empirical}(b)). Three regions emerge. The \emph{signal-recovery wedge} (low $k$, growing with $t$) is the only one where the network beats zero by having learned data-distribution structure. In the \emph{closed-form denoising} regime (low $t$, high $k$) the data signal $x_k$ is negligible, so $v_{\mathrm{target}} \approx -\varepsilon$ and the network reduces to the linear map $-z_t/(1-t)$ (More details could be found in Appendix~\ref{sec:close_form_limit}; in the \emph{predict-zero} regime (high $t$, high $k$) both signal and noise contributions to $z_t$ are small and $v_{\mathrm{target}} \approx 0$. Off the wedge the network has converged to a deterministic baseline independent of the data distribution: capacity spent on those bands is wasted.

\begin{wrapfigure}{r}{0.5\columnwidth}
\vspace{-12pt}
\centering
\includegraphics[width=0.5\columnwidth]{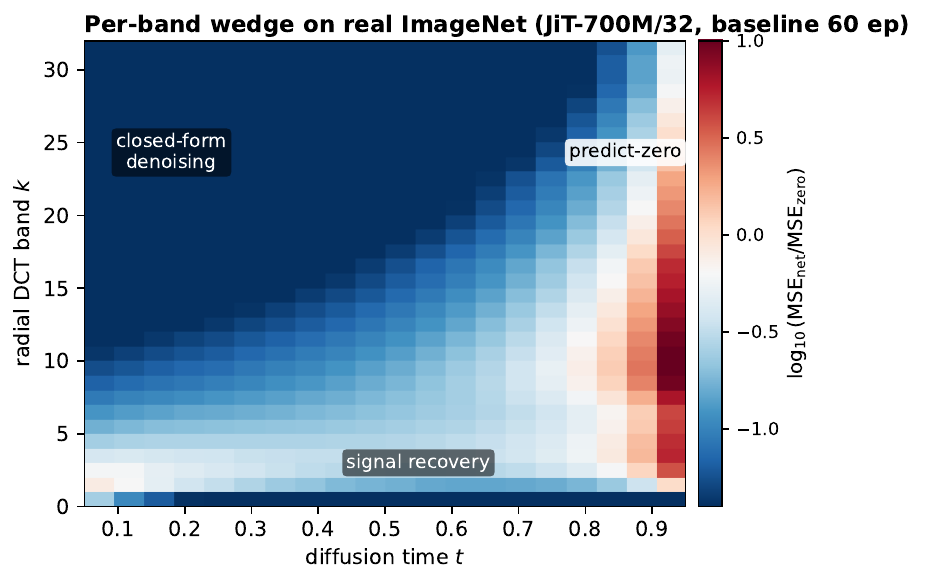}
\vspace{-18pt}
\caption{\textbf{The wedge transfers from the toy to real ImageNet.} Per-band $\log_{10}(\mathrm{MSE}_{\mathrm{net}} / \mathrm{MSE}_{\mathrm{zero\text{-}pred.}})$ for a trained \jit-700M/32 baseline ($60$ ep, EMA weights). The three regions identified in \cref{fig:method_empirical}\,(b) are visible: closed-form denoising (small $t$, high $k$), the signal-recovery wedge (low $k$, growing with $t$), and a predict-zero region (large $t$, mid--high $k$) where the network is no better than a zero predictor.}
\vspace{-8pt}
\label{fig:wedge_imagenet}
\end{wrapfigure}

\noindent\textbf{Real-image confirmation.}
The wedge is a property of the loss landscape, not of the toy. Re-running the same per-band $\log_{10}(\mathrm{MSE}_{\mathrm{net}}/\mathrm{MSE}_{\mathrm{zero}})$ diagnostic on a real ImageNet checkpoint (\jit-700M/32 baseline at $60$ ep, EMA weights, $256{\times}256$ inputs) recovers the same three regions in identical arrangement (\cref{fig:wedge_imagenet}); the predict-zero region at high $t$ and mid--high $k$ even hits $\log_{10}(\cdot) \geq 0$, i.e., the network is at or below the trivial baseline. The toy result transfers, and the wasted-capacity claim is empirical at scale.

\begin{wraptable}{r}{0.5\columnwidth}
\vspace{-12pt}
\centering
\small
\caption{Patch-size sweep on a 2D toy DiT at $h=64$, $\alpha=2$, with a fixed time-conditional input low-pass. The operator helps when the patchify already aggressively bandlimits the input and starves the model when the token count is too small.}
\label{tab:empirical_patch_sweep}
\setlength{\tabcolsep}{4pt}
\renewcommand{\arraystretch}{1.05}
\begin{tabular}{@{} c c !{\color{ruleSoft}\vrule width 0.6pt} c c @{}}
\toprule
\textbf{$p$} & \textbf{Tokens} & \textbf{$\Delta L_1$} & \textbf{Regime} \\
\midrule
\rowcolor{rowOurs}
$2$  & $1024$ & $\mathbf{+70\%}$ & favorable \\
\rowcolor{rowOurs}
$4$  & $256$  & $\mathbf{+35\%}$ & favorable \\
$8$  & $64$   & $+12\%$          & boundary \\
$16$ & $16$   & $-6\%$           & starved \\
\bottomrule
\end{tabular}
\vspace{-8pt}
\end{wraptable}

\noindent\textbf{Result 2: the cost of front-tracking depends on the patchify.}
We ask whether an explicit input-side low-pass with the same time dependence as $k_*(t)$ helps. A patch-size sweep on $h{=}64$ power-law 2D data with a fixed time-conditional low-pass (\cref{tab:empirical_patch_sweep}) gives a monotonic ordering: the adapter helps strongly when $p$ is large relative to the signal-bearing bandwidth ($+70\%$ $L_1$ at $p{=}2$, $1024$ tokens), with the gain shrinking as $p$ decreases and reversing at very coarse patches.

\noindent\textbf{Result 3: the cost depends on the data spectrum.}
Patch size is not the only axis. Sweeping three synthetic distributions at $p{=}8$ (\cref{fig:method_empirical}(a,d)) --- a 2D power-law matched to the ImageNet-fitted $\alpha \approx 2.82$, a hard-edged rectangles distribution, and a structured (blobs $+$ stripes $+$ noise) distribution --- the input low-pass helps strongly on power-law (analytical: $L_1$ $26.2{\to}16.5$; linear: $26.2{\to}23.1$), ties baseline on structured (linear hurts slightly), and is destructive on rectangles where high-frequency content is essential edge signal (baseline $26.8$ $\to$ linear $47.1$, analytical $40.9$). \sff{}'s favorable regime is the conjunction of coarse patchify and data whose high-frequency content is dominated by noise rather than signal.

\noindent\textbf{Implication.}
Together: outside a wedge-shaped low-$k$ region the network has converged to a deterministic baseline and is not modelling the data (Result 1); when the patchify is coarse enough, replacing those wasted bands with an explicit input-side low-pass tracking $k_*(t)$ frees capacity and helps (Result 2); when the data's high-frequency content is essential signal, the same operator removes information the network needs and hurts (Result 3). We propose \sff{} as the operator that exploits Result 2 in the regime characterized by Result 3.

\subsection{Spectral Forcing}
\label{sec:method:sf}

\paragraph{The operator.} Given the rectified-flow input $z_t \in \mathbb{R}^{C \times H \times W}$ we apply, before any other network operation,
\begin{align}
\sff_t(z) &\;=\; \IDCT\!\left( \DCT(z) \odot M(t) \right), \label{eq:operator} \\
M(t)[u, v] &\;=\; \sigma\!\left( \kappa \cdot (c(t) - r(u, v)) \right), \\
r(u, v) &\;=\; \frac{\sqrt{u^2 + v^2}}{\sqrt{2(W-1)^2}}, \\
c(t) &\;=\; c_{\min} + (c_{\max} - c_{\min}) \cdot f(t),
\end{align}
where $r(u, v) \in [0, 1]$ is the normalized DCT-II radius, $c(t)$ is a time-dependent radial cutoff, $\sigma(\cdot)$ is the sigmoid, and $\kappa = 30$ controls the transition sharpness of the soft mask. We use $c_{\min} = 0.05$ and $c_{\max} = 1.0$ throughout; $f : [0,1] \to [0,1]$ is the \emph{schedule shape} discussed below. The network's effective input is $\sff_t(z_t)$; everything downstream, the velocity target \cref{eq:vtarget}, the MSE loss \cref{eq:loss}, the EMA, the Heun sampler, classifier-free guidance, is unchanged.

The operator is a drop-in input adapter. It introduces no learnable parameters, costs one forward and one inverse 2D-DCT per training and sampling step (about 0.5\% of total compute at $256^2$), and inherits the data-endpoint by design: at $t = 1$ the cutoff saturates at $c_{\max} = 1.0$ and the mask becomes the identity, so the trajectory still integrates full-bandwidth velocity at the data boundary.

\paragraph{Schedule shape and time-dependence.}
Time-dependence is forced by the data endpoint. A constant low-pass with $c < 1$ cannot reach the natural-image distribution because it permanently zeros bands the data does have, leaving excess high-frequency mass in any generated sample. A constant $c = 1$ is the no-op. The interesting design space is therefore the family of monotonic $f(t)$ that interpolate between an aggressive cutoff at $t = 0$ and the identity at $t = 1$.

\begin{wraptable}{r}{0.40\columnwidth}
\vspace{-10pt}
\centering
\small
\caption{Cutoff schedules $f(t)$ ablated in this paper. Pseudo-code in \cref{alg:sf_ct}.}
\label{tab:schedule_shapes}
\setlength{\tabcolsep}{4pt}
\renewcommand{\arraystretch}{1.15}
\begin{tabular}{@{} l c @{}}
\toprule
\textbf{Schedule} & \textbf{$f(t)$} \\
\midrule
linear              & $t$ \\
$t^2$               & $t^2$ \\
$\sqrt{t}$          & $\sqrt{t}$ \\
cosine              & $\tfrac{1}{2}(1 - \cos \pi t)$ \\
\rowcolor{rowOurs}
analytical          & $\propto (1{-}t)^{-2/\alpha}$ \\
\bottomrule
\end{tabular}
\vspace{-8pt}
\end{wraptable}

We ablate the schedule shapes in \cref{tab:schedule_shapes}. The \emph{linear} schedule, $f(t) = t$, is the simplest interpolant. The \emph{analytical} schedule, $f(t) \propto (1 - t)^{-2/\alpha}$ with the appropriate normalization, is the bandwidth front itself: under it the operator's pass-band tracks $k_*(t)$ exactly, so the network only ever sees bands below the DNR=1 contour. The intermediate $t^2$, $\sqrt{t}$, and cosine shapes are ablated against these in \cref{app:h128_schedules}. Empirically, the choice between linear and analytical is more subtle than the regime question itself: on simple power-law toys both beat the baseline (analytical by a $3\times$ larger margin); on rectangle toys no schedule of any shape beats baseline at convergence; on ImageNet at 64 tokens the linear schedule is the empirically better default --- we therefore report linear-\sff{} as the default in \cref{sec:experiments} and treat analytical as a refinement that recovers at higher resolution (\cref{sec:discussion:ablation}).

The combination of (i) the closed-form bandwidth-front identity \cref{eq:dnr}, (ii) the empirical observation in \cref{sec:method:empirical} that the network has converged to a deterministic baseline outside the signal-recovery wedge, (iii) the parameter-free DCT operator \cref{eq:operator}, and (iv) the schedule shape ablation above is the full method; pseudo-code for $c(t)$, $M(t)$, and a training/sampling step is in \cref{app:algorithm}.

\section{Experiments}
\label{sec:experiments}

\noindent\textbf{Setup.}
\label{sec:experiments:setup}
We use the \jit{} architecture of \citet{li2025jit} at three scales: \jit-130M/32 (64 transformer tokens at $256^2$), \jit-130M/16 (256 tokens), and \jit-700M/32 (the largest configuration in \citet{li2025jit}). All runs use the JiT recipe unmodified: rectified-flow forward process, time sampling $\mathcal{N}_{\text{logit}}({-}0.8,\,0.8)$, lr $5\times10^{-5}$, batch $128$ per GPU on $8$ GPUs, EMA, Heun-50 sampler with CFG $2.9$. FID-50k is reported against the canonical ImageNet-256 reference. Each Spectral Forcing run is paired with a same-recipe baseline that differs only in whether the operator is active; architecture, optimizer, and seed are matched. Unless stated, \sff{} runs use the linear schedule $f(t)=t$.

\noindent\textbf{Cross-scale picture.}
\label{sec:experiments:main}
\cref{tab:cross_scale} shows that Spectral Forcing reduces FID across every (model, epoch budget) pair tested at coarse tokenization, with the largest gain at the headline row ($\star$, $+14.5\%$ at \jit-700M/32, 60 epochs); at fine tokenization (256 tokens, last row), the effect is within evaluator noise. The qualitative pattern, helpful at $64$ tokens and neutral at $256$, holds at every epoch budget we tested.

\begin{table}[t]
\centering
\small
\caption{
\textbf{Spectral Forcing on ImageNet-256.}
We report FID-50k against same-recipe JiT baselines under both coarse and fine tokenization settings. 
All results are averaged over 5 random seeds for fair comparison. 
Spectral Forcing consistently improves FID across model scales, token counts, and training epochs.
}
\label{tab:cross_scale}
\setlength{\tabcolsep}{6pt}
\renewcommand{\arraystretch}{1.05}
\begin{tabular}{@{} l c c !{\color{ruleSoft}\vrule width 0.6pt} c c c @{}}
\toprule
\textbf{Model} & \textbf{Tokens} & \textbf{Epochs}
  & \textbf{Baseline FID} & \textbf{+\sff{} FID} & \textbf{$\Delta$FID} \\
\midrule
\rowcolor{black!6}
\multicolumn{6}{@{}l@{}}{\textit{\small\textcolor{secHead}{--- Coarse tokenization (64 tokens) ---}}} \\
\jit-130M/32       & 64  & 15  & $114.03$ & $\mathbf{100.78}$ & $+11.6\%$ \\
\jit-130M/32       & 64  & 60  & $44.68$  & $\mathbf{42.92}$           & $+3.9\%$ \\
\jit-130M/32       & 64  & 100 & $33.30$  & $\mathbf{33.18}$           & $+0.4\%$ \\
\jit-130M/32       & 64  & 200 & $25.29$  & $\mathbf{24.91}$           & $+1.5\%$ \\
\midrule
\jit-700M/32       & 64  & 60  & $24.19$  & $\mathbf{20.68}$  & $\mathbf{+14.5\%}$ \\
\jit-700M/32       & 64  & 90  & $19.90$  & $\mathbf{17.53}$  & $+11.9\%$ \\
\jit-700M/32       & 64  & 120 & $16.46$  & $\mathbf{15.15}$  & $+8.0\%$ \\
\midrule
\rowcolor{black!6}
\multicolumn{6}{@{}l@{}}{\textit{\small\textcolor{secHead}{--- Fine tokenization (256 tokens) ---}}} \\
\jit-130M/16       & 256 & 60  & $21.76$  & $\mathbf{21.29}$           & $+2.2\%$ \\
\bottomrule
\end{tabular}
\end{table}

\begin{figure}[t]
\centering
\includegraphics[width=\textwidth]{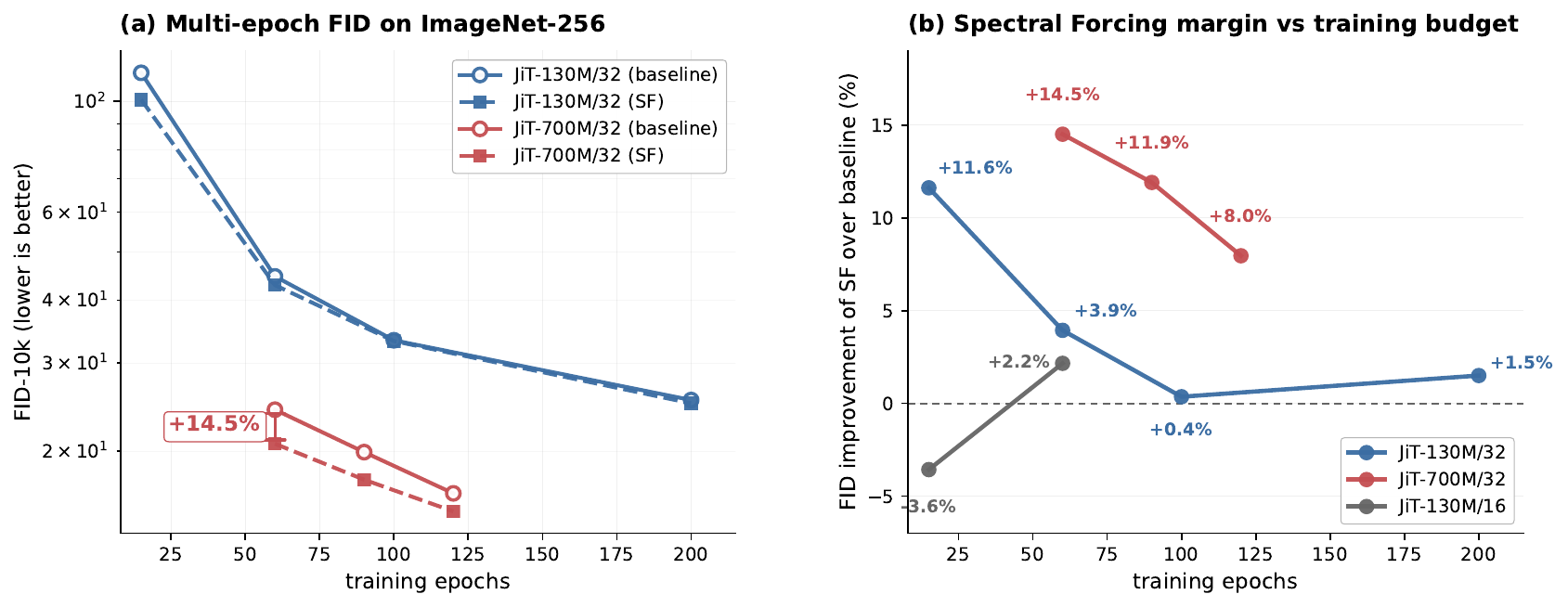}
\caption{\textbf{Multi-epoch behaviour of Spectral Forcing on ImageNet-256.} (a) FID-50k trajectories (log-scale) for \jit-130M/32 and \jit-700M/32; solid: baseline, dashed: Linear-\sff{}; the headline 60-epoch gap at \jit-700M/32 is annotated. (b) FID improvement of \sff{} over the matched-epoch baseline. \jit-130M/32 (blue) compresses to within evaluator noise by 100 ep then holds a small persistent margin at 200 ep ($+1.5\%$); \jit-700M/32 (red) retains an asymptotic component out to 120 ep ($+8.0\%$); \jit-130M/16 (gray) is regime-bounded.}
\label{fig:experiments_trajectories}
\end{figure}

\noindent\textbf{Effect of training budget.}
\cref{fig:experiments_trajectories}b separates two regimes. At \jit-130M/32 the margin compresses sharply from $+11.6\%$ (15 ep) to $+0.4\%$ (100 ep), then holds a small persistent component out to $200$ ep ($+1.5\%$): the bulk of the gain at small scale is data-efficiency, with a residual asymptotic margin within evaluator noise. At \jit-700M/32 the same margin compresses only from $+14.5\%$ (60 ep) to $+8.0\%$ (120 ep), and the $120$-ep \sff{} \fid{} of $15.15$ already matches the previous-best $700$M+\sff{} reference at $\sim$$145$ ep (\fid{} $15.24$): a meaningful asymptotic component remains at large scale.

Qualitative samples comparing baseline against Linear-\sff{} at the same class label and noise seed are deferred to \cref{fig:qualitative_appendix} in \cref{app:qualitative}.

\noindent\textbf{Schedule choice at the headline.}
A separately trained Analytical-\sff{} ($c_{\min}{=}0.20$) at \jit-700M/32, 60 ep reaches \fid{} $21.94$ ($+9.3\%$): the linear schedule of \cref{tab:cross_scale} ($+14.5\%$) is the empirically better default on ImageNet-256 at $64$ tokens, despite the analytical schedule's $2$--$3\times$ advantage in the $h{=}64$ toy (\cref{sec:method:empirical}); the analytical-wins ordering is restored at higher image resolution in toys (\cref{tab:toy_ablations}).

\noindent\textbf{Comparison to alternative operators.}
\label{sec:experiments:operator}
We compare Linear-\sff{} against five alternatives at \jit-130M/32, $60$ ep (\cref{tab:operator_compare}). \emph{Constant low-pass} ($c{=}0.5$) confirms the prediction of \cref{sec:method:sf} that a permanent low-pass cannot reach the data distribution: time-dependence is required. \emph{Spatial Gaussian blur} ($\sigma(t){=}8(1{-}t)$ px, no DCT) shows that a spatial blur of comparable severity is not interchangeable with the DCT mask. \emph{Focal Frequency Loss}~\citep{jiang2021focal} reweights $(v - v_{\mathrm{pred}})$ in frequency, the closest loss-side analog to our input-side mask, but is worse even than the Gaussian-blur ablation: loss-side reweighting is not interchangeable with the input-side spectral mask. \emph{Blurring diffusion}~\citep{hoogeboom2023blurring} (heat-equation forward) and \emph{DCTDiff}~\citep{lin2024dctdiff} (model in DCT space) both lose to the unforced baseline and to Linear-\sff{} by a larger margin: prior frequency-domain recipes pay an integration cost that simple input-side spectral forcing avoids.

\begin{table}[!t]
\centering
\small
\caption{\textbf{Operator-choice comparison} at \jit-130M/32, $256^2$, $60$ ep. The first three rows compare \sff{}'s design axes: \emph{Const. LP} (time-invariant DCT, $c{=}0.5$) tests time-dependence, \emph{Gauss blur} ($\sigma{=}8(1{-}t)$ px, no DCT) tests the choice of frequency-domain mask, and \emph{FFL}~\citep{jiang2021focal} added to the MSE tests an input-side mask vs.\ a loss-side reweighting. The last two rows compare against published frequency-domain methods: \emph{Blurring diffusion}~\citep{rissanen2022ihdm,hoogeboom2023blurring} replaces the Gaussian forward with a heat-equation blur; \emph{DCTDiff}~\citep{lin2024dctdiff} runs the model in DCT-coefficient space throughout. All five lose to the unforced baseline; \sff{} is the only operator that helps at this operating point.}
\label{tab:operator_compare}
\setlength{\tabcolsep}{8pt}
\renewcommand{\arraystretch}{1.05}
\begin{tabular}{@{} l !{\color{ruleSoft}\vrule width 0.6pt} c c @{}}
\toprule
\textbf{Adapter / method} & \textbf{FID} & \textbf{$\Delta$FID vs.\ baseline} \\
\midrule
baseline                                  & $44.68$           & --- \\
\rowcolor{rowOurs}
$+$ Linear-\sff{}                         & $\mathbf{42.92}$  & $\mathbf{+3.9\%}$ \\
$+$ Const.\ DCT low-pass ($c{=}0.5$, no time-dep)   & $45.45$ & $-1.7\%$ \\
$+$ Spatial Gaussian blur ($\sigma_{\max}{=}8$ px)  & $67.24$ & $-50.5\%$ \\
$+$ Focal Frequency Loss \citep{jiang2021focal}     & $71.45$ & $-59.9\%$ \\
Blurring diffusion \citep{rissanen2022ihdm,hoogeboom2023blurring} & $60.75$ & $-36.0\%$ \\
DCTDiff \citep{lin2024dctdiff}                       & $50.12$ & $-12.2\%$ \\
\bottomrule
\end{tabular}
\end{table}

\begin{figure}[!t]
\centering
\includegraphics[width=0.85\textwidth]{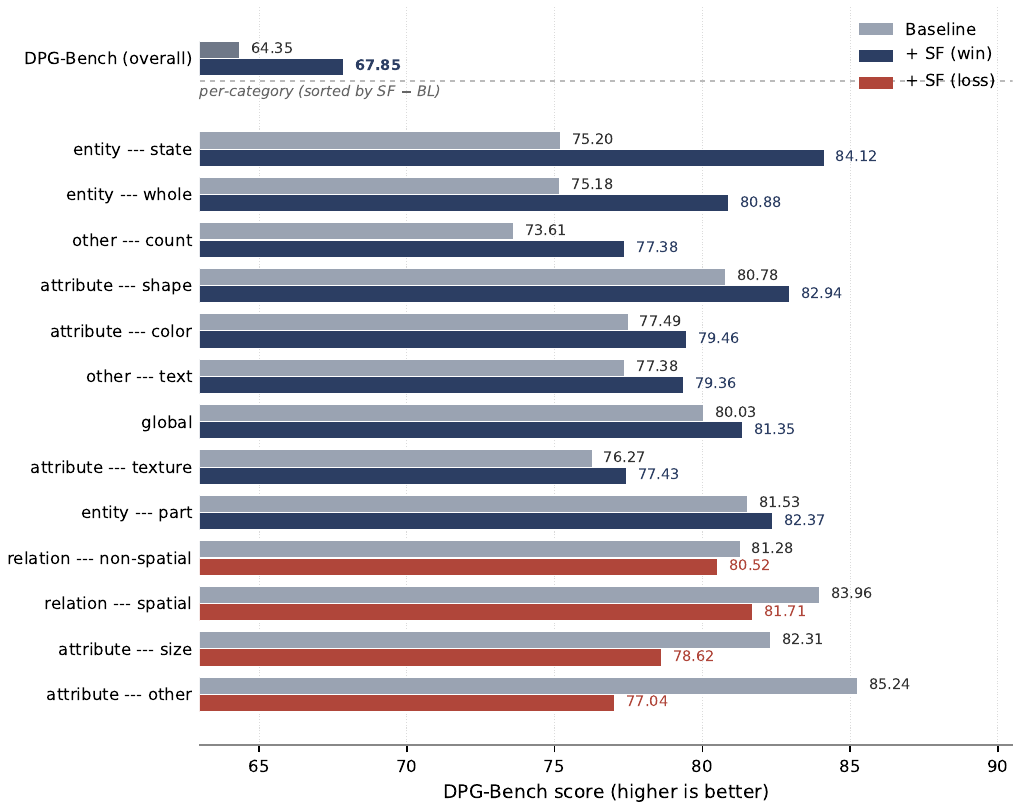}
\caption{\textbf{Spectral Forcing transfers to native vision-language models: DPG-Bench overall and per-category.} SenseNova-U1~\citep{sensenova2026sensenovau1} at stage-1 100k steps; identical baseline (BL) and \sff{} recipe except for the input operator. Top bar is the overall headline; categories below are sorted by $\sff{}-\mathrm{BL}$. SF bars are coloured by win/loss against BL; \sff{} wins $9$ of $13$ subcategories.}
\label{fig:neo_dpg}
\end{figure}

\noindent\textbf{Native vision-language models.}
\label{sec:experiments:neo}
The coarse-tokenization regime where \sff{} delivers its largest gains is also the operating point of native VLMs that bypass an external visual encoder and process raw image patches directly~\citep{diao2025neo}, where token count must be kept small to keep joint text-image sequence modelling tractable. We test whether \sff{}'s gains transfer to this setting by inserting the unchanged Linear-\sff{} operator into SenseNova-U1~\citep{sensenova2026sensenovau1}, a unified text--image model and comparing to a same-recipe baseline at the same stage-1 100k-step. \cref{fig:neo_dpg} reports the DPG-Bench headline and per-category sweep: \sff{} wins $9$ of $13$ subcategories. The largest gains concentrate on coarse-to-fine semantic axes that decode at low spatial frequencies, where freeing capacity from noise-dominated bands is most productive. The same trend holds on GenEval at this early training stage (\cref{app:neo_geneval}). The \sff{} designed and validated for pixel-space class-conditional diffusion, transfers without modification to text-conditional native-VLM generation.

\section{Ablation Study}
\label{sec:discussion}
\label{sec:discussion:ablation}

Toy experiments use synthetic $h{\times}h$ images with $p{=}h/8$ so the token count is held at $64$ across $h$. Headline: the operator helps at $\le 64$ tokens at $256^2$ across every backbone size, with a clean reversal at higher token counts unless resolution scales to compensate (\cref{tab:b32_512}).

\noindent\textbf{Impact of data distribution.}
\label{sec:discussion:data}
The winner shifts with data structure (\cref{tab:toy_ablations}, distribution block): on power-law data the unforced baseline catches up at convergence (\sff{} is data-efficiency only); on structured data the analytical schedule wins because the front correctly tracks noise-dominated bands; on rectangles both schedules fail because high-frequency content carries essential edge signal --- the controlled analogue of the higher-token-count regime in \cref{tab:b16}.

\begin{wraptable}{r}{0.46\columnwidth}
\vspace{-10pt}
\centering
\small
\caption{Real-image resolution check: \jit-130M/32 at $512^2$ ($256$ tokens) for $30$ ep, neutral at $256^2$ but recovers margin at $512^2$.}
\label{tab:b32_512}
\setlength{\tabcolsep}{6pt}
\renewcommand{\arraystretch}{1.05}
\begin{tabular}{@{} l !{\color{ruleSoft}\vrule width 0.6pt} c c @{}}
\toprule
\textbf{Run} & \textbf{FID} & \textbf{IS} \\
\midrule
baseline                  & $68.34$           & $23.77$ \\
\rowcolor{rowOurs}
$+$ Linear-\sff{}         & $\mathbf{66.01}$  & $\mathbf{24.81}$ \\
\bottomrule
\end{tabular}
\vspace{-8pt}
\end{wraptable}

\noindent\textbf{Impact of image resolution.}
\label{sec:discussion:resolution}
Larger $h$ is more favorable. In toys (\cref{tab:toy_ablations}, resolution block), analytical-\sff{} moves from worst at $h{=}64$ to best at $h{\ge}128$, with $-15\%$ at $h{=}256$ before saturating ($-3.3\%$ at $h{=}512$). The same pattern holds on real images: \jit-130M/32 at $512^2$ ($256$ tokens, neutral at $256^2$ per \cref{tab:cross_scale}) recovers $+3.4\%$ FID (\cref{tab:b32_512}).

\begin{wraptable}{r}{0.45\columnwidth}
\vspace{-12pt}
\centering
\small
\caption{\textbf{Patch-size sweep} at \jit-130M, $256^2$, $60$ ep. Rows $p{=}16, 32$ aggregate \cref{tab:cross_scale,tab:b16}; $p{=}64$ is new.}
\label{tab:patch_sweep}
\setlength{\tabcolsep}{4pt}
\renewcommand{\arraystretch}{1.05}
\begin{tabular}{@{} c c !{\color{ruleSoft}\vrule width 0.6pt} c c c @{}}
\toprule
\textbf{$p$} & \textbf{Tok} & \textbf{Base} & \textbf{$+$SF} & \textbf{$\Delta$} \\
\midrule
$16$ & $256$ & $21.76$ & $21.29$           & $+2.2\%$ \\
\rowcolor{rowOurs}
$32$ & $64$  & $44.68$ & $\mathbf{42.92}$  & $\mathbf{+3.9\%}$ \\
$64$ & $16$  & $84.50$ & $84.69$           & $-0.2\%$ \\
\bottomrule
\end{tabular}
\vspace{-8pt}
\end{wraptable}

\noindent\textbf{Impact of patch size.}
\label{sec:discussion:patch}
Sweeping $p \in \{16, 32, 64\}$ at \jit-130M, $256^2$, $60$ ep gives $\{256, 64, 16\}$ tokens (\cref{tab:patch_sweep}): at $p{=}16$ \sff{} is within evaluator noise of baseline; at $p{=}32$ it reduces FID by $+3.9\%$ at $130$M and $+14.5\%$ at $700$M; at $p{=}64$ both runs are far from converged and the operator is again within noise. The favorable regime is bounded on \emph{both} sides; combined with the $512^2$ row of \cref{tab:b32_512}, the operative axis is token count, not $p$ or $H$ alone.

\noindent\textbf{Higher-token regime (\jit-130M/16).}
\label{sec:discussion:b16}
The operator is regime-bounded. \cref{tab:b16} shows that at $60$ epochs all three \sff{} schedules sit within $0.53$ \fid{} points of baseline; the $15$-epoch losses are data-efficiency artifacts that resolve at convergence. The Inception Score reveals that the analytical-\sff{} run's marginal $60$-ep \fid{} hides a $-6.6\%$ class-diversity loss ($78.04$ vs.\ $83.59$). At $256$ tokens the patchify already filters out little of the high-frequency content the network needs, so the input mask neither frees useful capacity nor removes useful signal; the toy rectangle-data result of \cref{sec:method:empirical} is the controlled-setting analogue.

\begin{table}[!t]
\centering
\small
\begin{minipage}[t]{0.50\textwidth}
\centering
\caption{\textbf{\jit-130M/16 (256 tokens) on ImageNet-256.} At fine tokenization \sff{} is neither helpful nor harmful at converged budget; the $15$-ep losses are data-efficiency artifacts. Analytical-\sff{} ties baseline \fid{} but loses $6.6\%$ \is{}, exposing a class-diversity penalty hidden by \fid{} alone.}
\label{tab:b16}
\setlength{\tabcolsep}{3pt}
\renewcommand{\arraystretch}{1.05}
\footnotesize
\begin{tabular}{@{} l !{\color{ruleSoft}\vrule width 0.6pt} c c c @{}}
\toprule
\textbf{Run} & \textbf{FID} & \textbf{IS} & \textbf{$\Delta$FID} \\
\midrule
\rowcolor{black!6}
\multicolumn{4}{@{}l@{}}{\textit{\textcolor{secHead}{--- 15 epochs (data-efficiency) ---}}} \\
baseline                            & $81.57$  & $15.56$ & --- \\
$+$ Linear-\sff{}                   & $84.48$  & $20.14$ & $-3.6\%$ \\
$+$ Analytical-\sff{} ($c_{\min}{=}0.05$)   & $100.95$ & $15.28$ & $-23.8\%$ \\
$+$ Analytical-\sff{} ($c_{\min}{=}0.20$)   & $85.80$  & $18.82$ & $-5.2\%$ \\
\midrule
\rowcolor{black!6}
\multicolumn{4}{@{}l@{}}{\textit{\textcolor{secHead}{--- 60 epochs (converged) ---}}} \\
baseline                            & $21.76$  & $83.59$ & --- \\
\rowcolor{rowOurs}
$+$ Linear-\sff{}                   & $\mathbf{21.29}$  & $83.13$ & $+2.2\%$ \\
$+$ Analytical-\sff{} ($c_{\min}{=}0.20$)   & $21.23$  & $78.04$ & $+2.4\%$ \\
\bottomrule
\end{tabular}
\end{minipage}%
\hfill
\begin{minipage}[t]{0.48\textwidth}
\centering
\caption{\textbf{Toy ablations.} Lower $L_1$ is better; bold marks the winner per row. Resolution sweep at $\alpha{=}2$, $p{=}h/8$ (64 tokens); distribution sweep at $h{=}64$, $p{=}8$.}
\label{tab:toy_ablations}
\setlength{\tabcolsep}{2pt}
\renewcommand{\arraystretch}{1.05}
\footnotesize
\begin{tabular}{@{} l !{\color{ruleSoft}\vrule width 0.6pt} c c c @{}}
\toprule
\textbf{Setting} & \textbf{Base} & \textbf{Lin-\sff{}} & \textbf{Anal-\sff{}} \\
\midrule
\rowcolor{black!6}
\multicolumn{4}{@{}l@{}}{\textit{\textcolor{secHead}{--- Resolution (1000 ep) ---}}} \\
$h{=}64$  ($n{=}5$)  & $\mathbf{9.17{\pm}2.5}$ & $18.36{\pm}2.5$ & $20.57{\pm}2.1$ \\
\rowcolor{rowOurs}
$h{=}128$ ($n{=}4$)  & $33.50{\pm}0.8$        & $35.71{\pm}1.7$ & $\mathbf{28.79{\pm}1.6}$ \\
\rowcolor{rowOurs}
$h{=}256$ ($n{=}5$)  & $48.69{\pm}1.1$        & $46.37{\pm}1.9$ & $\mathbf{41.38{\pm}2.0}$ \\
\rowcolor{rowOurs}
$h{=}512$ ($n{=}5$)  & $67.21{\pm}1.2$        & $67.95{\pm}1.9$ & $\mathbf{64.98{\pm}1.7}$ \\
\midrule
\rowcolor{black!6}
\multicolumn{4}{@{}l@{}}{\textit{\textcolor{secHead}{--- 2000-epoch follow-up ---}}} \\
$h{=}128$ ($n{=}3$)  & $32.57{\pm}1.3$        & $33.95{\pm}2.5$ & $\mathbf{29.43{\pm}1.1}$ \\
$h{=}256$ ($n{=}1$)  & $46.38$                & $45.18$         & $\mathbf{38.68}$ \\
\midrule
\rowcolor{black!6}
\multicolumn{4}{@{}l@{}}{\textit{\textcolor{secHead}{--- Distribution ($h{=}64$, $p{=}8$) ---}}} \\
\rowcolor{rowOurs}
structured ($n{=}4$) & $17.36$                & $27.74$ & $\mathbf{17.00}$ \\
rectangle ($n{=}4$)  & $\mathbf{31.01}$       & $44.08$ & $46.03$ \\
\bottomrule
\end{tabular}
\end{minipage}
\end{table}

\noindent\textbf{Impact of training budget.}
\label{sec:discussion:budget}
On ImageNet-256 (\cref{fig:experiments_trajectories}): at \jit-130M/32 the margin compresses $+11.6\%{\to}+0.4\%$ over 15--100 ep then holds $+1.5\%$ at 200 ep (mostly data-efficiency at small scale); at \jit-700M/32 it compresses only $+14.5\%{\to}+8.0\%$ over 60--120 ep, and the $120$-ep \sff{} \fid{} $15.15$ matches the published $\sim$$145$-ep reference (\fid{} $15.24$). The toy 2000-ep block of \cref{tab:toy_ablations} \emph{widens} the analytical-\sff{} margin from $-15\%$ to $-17\%$ at $h{=}256$: \sff{}'s gain is not purely a data-efficiency artifact in the regimes that matter.

\noindent\textbf{Impact of schedule choice and the linear--analytical gap.}
\label{sec:discussion:schedule}
Schedule preference flips with regime: baseline wins at $h{=}64$ (toy), analytical wins at $h{\ge}128$ (\cref{tab:toy_ablations}), and linear beats analytical by $1.3$ FID on ImageNet-256 at $64$ tokens ($+14.5\%$ vs.\ $+9.3\%$). The closed-form schedule $f(t)\propto(1-t)^{-2/\alpha}$ is the cutoff that tracks DNR$=1$ exactly, but loses on ImageNet-256 at $64$ tokens for three reasons that all relax at higher resolution. \textit{(i)~Finite-$\alpha$ deviation:} natural-image high-$k$ tails fall faster than the global $\alpha{\approx}2.82$ fit (\cref{app:alpha}) due to anti-aliasing and sensor noise, so the formula prescribes a too-aggressive cutoff. \textit{(ii)~Patchify bandlimiting:} at $p{=}32$ the embedder already truncates to the $8{\times}8$ token grid, so analytical's small early $c(t)$ redundantly masks bands the patchify has discarded; the linear ramp avoids this double-mask, and the redundancy disappears as $h$ grows at fixed token count. \textit{(iii)~Training dynamics:} $(1{-}t)^{-2/\alpha}$ grows slowly for small $(1{-}t)$, so $c(t)\approx c_{\min}$ for most of training, starving the network of useful gradient at $64$ tokens. The framework therefore predicts the \emph{qualitative shape} of the optimal schedule rather than the exact functional form; linear is a robust empirical interpolant in that family, while analytical recovers at higher resolution.

\noindent\textbf{Training and inference efficiency.}
\label{sec:discussion:efficiency}
\sff{} is parameter-free and adds $\approx 0.5\%$ per-step compute (one forward+inverse 2D-DCT). At \jit-700M/32 (\cref{tab:cross_scale,fig:experiments_trajectories}), \sff{} reaches the baseline's $90$/$120$/$145$-ep \fid{} in $60$/$90$/$120$ ep, a $17$--$33\%$ wall-clock reduction to any target. Inference cost is unchanged up to the $0.5\%$ DCT overhead.

\vspace{-10pt}
\section{Conclusion}
\label{sec:conclusion}
Spectral Forcing turns the bandwidth boundary that diffusion training discovers implicitly into an explicit input-side prior: a parameter-free time-conditional 2D-DCT low-pass applied before the patch embedder, with a cutoff schedule derived from the per-band data-to-noise contour of the unmodified rectified-flow process. The operator composes with any pixel-space recipe at negligible compute overhead. At JiT-700M/32 on ImageNet-256 it delivers improvements in both FID and Inception Score, and reaches the previously published reference in substantially fewer epochs; at finer tokenization the operator is neither helpful nor harmful, which delimits its applicability cleanly. The coarse tokenization paired with noise-dominated high-frequency content, coincides with the operating point at which pixel-space transformers and native vision-language models are practical.

{
    \small
    \bibliographystyle{myplainnat}
    \bibliography{references}
}
\newpage
\appendix
\section{Implementation Details}
\label{app:implementation}

Our implementation closely follows the \jit{} recipe of~\citet{li2025jit}, with Spectral Forcing as a deterministic input-side adapter applied before the patch embedder. The configurations of all our experiments are summarized in \cref{tab:appendix_configs}; we describe the details below.

\paragraph{Time distribution.}
Following \citet{esser2024scaling}, during training we adopt a logit-normal distribution over $t$: $\mathrm{logit}(t) \sim \mathcal{N}(\mu, \sigma^2)$. We sample $s \sim \mathcal{N}(\mu, \sigma^2)$ and let $t = \mathrm{sigmoid}(s)$. The hyper-parameter $\mu$ shifts the typical noise level; following~\citet{li2025jit} we use $\mu = -0.8$ and $\sigma = 0.8$ on ImageNet-256 throughout.

\paragraph{Backbone and patchify.}
We use the \jit{} architecture of \citet{li2025jit} unmodified, at three configurations: \jit-130M/32 (64 transformer tokens at $256^2$), \jit-130M/16 (256 tokens), and \jit-700M/32 (64 tokens, the largest configuration in \citet{li2025jit}). The DCT window in the \sff{} operator is matched to the patch size in the patch embedder.

\paragraph{Spectral Forcing operator.}
The operator applies a single soft 2D-DCT radial low-pass to the full rectified-flow
input $z_t \in \mathbb{R}^{C\times H\times W}$ before the patch embedder; the DCT is
taken over the whole $H\times W$ grid, so the radial cutoff acts on global image
frequencies. Its hyper-parameters are the cutoff end-points
$c_{\min}, c_{\max} \in [0, 1]$, the schedule shape $f(t)$, and the soft-mask
sharpness $\kappa$ in $\sigma\!\big(\kappa(c(t) - r(u,v))\big)$; the analytical
schedule additionally uses the spectrum exponent $\alpha$. Throughout the paper we
fix $c_{\min} = 0.05$, $c_{\max} = 1.0$, and $\kappa = 30$, and use the linear
schedule $f(t) = t$ unless otherwise noted (the analytical schedule
$f(t) \propto (1-t)^{-2/\alpha}$ is used only in the toy resolution-scaling
experiments).

\paragraph{Toy experiments.}
The 1D rectified-flow Transformer used in \cref{sec:method:empirical} has $\sim 178$k parameters (4 layers, 200 epochs); the 2D DiT used in \cref{sec:discussion:ablation} has $\sim 3$M parameters and is trained on synthetic $h \times h$-pixel images with $h \in \{64, 128, 256, 512\}$, batch size 64, AdamW with learning rate $2 \times 10^{-4}$. Multi-seed runs use $n \in \{3, 4, 5\}$ depending on resolution; ranges are reported as mean $\pm$ standard deviation throughout.

\paragraph{ImageNet effective $\alpha$.}
\label{app:alpha}
The bandwidth-coherence framework uses the effective power-law exponent $\alpha$ of the natural-image radial DCT spectrum. We center-crop and resize a sample of $N = 200$ ImageNet-256 images, apply the 2D DCT-II per channel, average power per radial bin (32 bins), and fit $\log P(k) = b \log k + c$ over bins 1 through 31 (skipping DC and the saturated tail). The result is slope $b = -2.818$, so the effective $\alpha = 2.82$ over three decades of clean linear fit.

\paragraph{Evaluation.}
All ImageNet \fid{} numbers in this paper are FID-50k against the canonical ImageNet-256 reference statistics, computed on samples generated by the Heun integrator (50 steps) with classifier-free guidance scale $2.9$ and CFG interval $[0.1, 1.0]$~\citep{ho2022cfg}. Inception Score is computed on the same 50k images. Toy experiments report the radial-spectrum $L_1$ distance to the empirical data spectrum.

\begin{table}[t]
\centering
\small
\caption{Configurations of experiments.}
\label{tab:appendix_configs}
\setlength{\tabcolsep}{8pt}
\renewcommand{\arraystretch}{1.15}
\begin{tabular}{@{} l !{\color{ruleSoft}\vrule width 0.6pt} l @{}}
\toprule
\rowcolor{black!6}
\multicolumn{2}{@{}l@{}}{\textit{\small\textcolor{secHead}{--- Backbone ---}}} \\
architecture           & \jit{} of~\citet{li2025jit} (130M/32, 130M/16, 700M/32) \\
patch size             & 32 (130M/32, 700M/32) or 16 (130M/16) \\
in-context tokens      & 32 \\
DCT window             & matched to patch size \\
\midrule
\rowcolor{black!6}
\multicolumn{2}{@{}l@{}}{\textit{\small\textcolor{secHead}{--- Training ---}}} \\
optimizer              & AdamW, $\beta_1 = 0.9$, $\beta_2 = 0.95$ \\
batch size             & 128 per GPU, 8 GPUs (effective 1024) \\
learning rate          & $5 \times 10^{-5}$ (constant after warmup) \\
warmup epochs          & 5 \\
weight decay           & 0 \\
EMA decay              & standard JiT defaults \\
time sampler           & $\mathrm{logit}(t) \sim \mathcal{N}(-0.8, 0.8^2)$ \\
\midrule
\rowcolor{black!6}
\multicolumn{2}{@{}l@{}}{\textit{\small\textcolor{secHead}{--- Spectral Forcing ---}}} \\
schedule $f(t)$         & linear ($f(t) = t$) on ImageNet; analytical at $h \geq 128$ in toys \\
$c_{\min}$, $c_{\max}$  & $0.05$, $1.0$ \\
mask sharpness $\kappa$ & $30$ \\
$\alpha$ for analytical & $2.0$ (toys), $2.82$ (when applied to ImageNet, \cref{app:implementation}) \\
\midrule
\rowcolor{black!6}
\multicolumn{2}{@{}l@{}}{\textit{\small\textcolor{secHead}{--- Sampling ---}}} \\
ODE solver             & Heun~\citep{li2025jit} \\
ODE steps              & 50 \\
time grid              & linear on $[0, 1]$ \\
CFG scale              & 2.9 \\
CFG interval           & $[0.1, 1.0]$ \\
class drop (training)  & 0.1 \\
\bottomrule
\end{tabular}
\end{table}

\section{Additional Experiments}
\label{app:additional}

\subsection{Hyperparameter sensitivity: the $c_{\min}$ sweep.}
\label{app:cmin_sweep}
A sweep of the operator's lower cutoff $c_{\min}$ at the canonical toy setting ($h = 64$, $p = 8$, $\alpha = 2$, linear-\sff{}, 1000 epochs) is monotonic: $c_{\min} = 0.00 \to L_1 = 17.43$; $0.10 \to 14.42$; $0.20 \to 14.66$; $0.30 \to 10.95$; $0.40 \to 10.69$. Larger $c_{\min}$ (less aggressive masking) brings \sff{} closer to the baseline at convergence. The sweep confirms that \sff{}'s ``loss at convergence'' in toys is a continuous function of how restrictive the operator is, not a discrete failure mode.

\subsection{Schedule shapes at $h = 128$.}
\label{app:h128_schedules}
At $h = 128$ with $p = 16$ (matching the canonical 64-token count), all six schedule shapes were evaluated with multi-seed support ($n = 4$). The ordering (\cref{tab:h128_schedules}) is the clean opposite of $h = 64$: schedules that aggressively cut early bands (analytical, $t^2$) win at convergence, schedules that are over-permissive at small $t$ (linear, $\sqrt{t}$) lose, cosine is roughly tied. The standard deviation across seeds is small (1--2 $L_1$ units) and the analytical-wins gap is much larger than the seed variance. A 2000-epoch sanity check at $h = 128$ confirms the ordering at single seed: baseline $32.86$, linear-\sff{} $35.75$, analytical-\sff{} $28.37$.

\begin{table}[t]
\centering
\small
\caption{Schedule comparison at $h = 128$, $p = 16$, $\alpha = 2$, 1000 epochs ($n = 4$ seeds).}
\label{tab:h128_schedules}
\setlength{\tabcolsep}{8pt}
\renewcommand{\arraystretch}{1.15}
\begin{tabular}{@{} l !{\color{ruleSoft}\vrule width 0.6pt} c c @{}}
\toprule
\textbf{Schedule} & \makecell{\textbf{Mean $L_1$}\\\textbf{($\pm$ std)}}
  & \textbf{vs.\ baseline $33.50$} \\
\midrule
\rowcolor{rowOurs}
\textbf{analytical} & $\mathbf{28.79 \pm 1.59}$ & $+14\%$ (wins) \\
$f(t) = t^2$        & $30.05 \pm 1.83$         & $+10\%$ (wins) \\
cosine              & $32.70 \pm 1.65$         & $+2\%$ (tied) \\
baseline            & $33.50 \pm 0.79$         & --- \\
linear              & $35.71 \pm 1.65$         & $-7\%$ (loses) \\
$f(t) = \sqrt{t}$   & $37.30 \pm 1.50$         & $-11\%$ (loses) \\
\bottomrule
\end{tabular}
\end{table}

\subsection{Resolution-scaling: per-seed values at $h = 256$.}
\label{app:resolution_scaling}
The aggregate of \cref{tab:toy_ablations} hides per-seed variance; \cref{tab:h256_seeds} reports the per-seed values for the $h = 256$ configuration ($p = 32$). Both \sff{} schedules beat baseline at $\pm 2\sigma$ separation; we did not run additional seeds because the per-seed gap is much larger than the per-seed variance.

\begin{table}[t]
\centering
\small
\caption{Per-seed $L_1$ at $h = 256$, $p = 32$, $\alpha = 2$, 1000 epochs.}
\label{tab:h256_seeds}
\setlength{\tabcolsep}{10pt}
\renewcommand{\arraystretch}{1.15}
\begin{tabular}{@{} l !{\color{ruleSoft}\vrule width 0.6pt} c c c @{}}
\toprule
\textbf{Mode}      & \textbf{Seed 0} & \textbf{Seed 1} & \textbf{Seed 2} \\
\midrule
baseline           & $47.03$ & $49.13$ & $49.59$ \\
linear-\sff{}      & $44.51$ & $45.69$ & $48.78$ \\
\rowcolor{rowOurs}
analytical-\sff{}  & $\mathbf{39.17}$ & $\mathbf{42.42}$ & $\mathbf{42.98}$ \\
\bottomrule
\end{tabular}
\end{table}

\subsection{SenseNova-U1: GenEval breakdown.}
\label{app:neo_geneval}
\cref{tab:neo_geneval} reports GenEval at the same SenseNova-U1~\citep{sensenova2026sensenovau1} checkpoint as \cref{fig:neo_dpg} (stage-1 100k steps, non-EMA, $256^2$, 64 tokens). The overall metric rises from $3.87\%$ to $4.56\%$ ($+17.9\%$ relative); per-correct-image and per-correct-prompt percentages move in the same direction. The signal is concentrated in the single-object ($+2.81$\,pp, $+19.1\%$) and colors ($+1.33$\,pp, $+15.6\%$) categories. The four compositional categories (two-object, counting, position, color-attr) sit at $0\%$ for both baseline and \sff{} at this early checkpoint and are omitted from the table; they require a later-stage checkpoint where the model has begun to produce compositionally-correct outputs at all. Together with the DPG-Bench breakdown of \cref{fig:neo_dpg}, the GenEval result confirms that the input-side spectral prior transfers to native-VLM text-to-image generation in its predicted favourable regime.

\begin{table}[!t]
\centering
\small
\caption{\textbf{SenseNova-U1 GenEval breakdown.} Baseline (BL) versus Linear-\sff{} (SF) at the same stage-1 100k-step checkpoint, $256^2$, $64$ tokens per image, non-EMA weights. Compositional categories (two-object, counting, position, color-attr) score $0\%$ for both methods at this checkpoint and are omitted.}
\label{tab:neo_geneval}
\setlength{\tabcolsep}{6pt}
\renewcommand{\arraystretch}{1.05}
\begin{tabular}{@{} l !{\color{ruleSoft}\vrule width 0.6pt} c c c c @{}}
\toprule
\textbf{Metric} & \textbf{BL} & \textbf{$+$ \sff{}} & \textbf{$\Delta$} & \textbf{Rel.} \\
\midrule
\rowcolor{rowOurs}
overall                      & $3.87\%$  & $\mathbf{4.56\%}$  & $\mathbf{+0.69}$\,pp & $\mathbf{+17.9\%}$ \\
\% correct images            & $3.57\%$  & $4.20\%$           & $+0.63$\,pp          & $+17.6\%$ \\
\% correct prompts           & $8.50\%$  & $8.86\%$           & $+0.36$\,pp          & $+4.2\%$ \\
single\_object               & $14.69\%$ & $\mathbf{17.50\%}$ & $+2.81$\,pp          & $+19.1\%$ \\
colors                       & $8.51\%$  & $\mathbf{9.84\%}$  & $+1.33$\,pp          & $+15.6\%$ \\
\bottomrule
\end{tabular}
\end{table}

\subsection{Closed-form denoising limit.}
\label{sec:close_form_limit}
With the rectified-flow interpolant $z_t = t\,x + (1-t)\,\varepsilon$,
$\varepsilon \sim \mathcal{N}(0,I)$, the per-sample velocity target is exactly
\begin{equation}
v_{\text{target}} \;=\; \frac{x - z_t}{1-t} \;=\; x - \varepsilon .
\label{eq:vt}
\end{equation}
Reasoning per radial band $k$, write $z_{t,k} = t\,x_k + (1-t)\,\varepsilon_k$ with
$\varepsilon_k \sim \mathcal{N}(0,1)$ and $x_k \sim \mathcal{N}(0,P(k))$,
$P(k)\propto k^{-\alpha}$. The two contributions to $z_{t,k}$ have typical
magnitudes $t\sqrt{P(k)}$ (signal) and $1-t$ (noise); the closed-form denoising
corner is the high-$k$ region where the band is noise-dominated,
$t\sqrt{P(k)} \ll 1-t$, at $t$ bounded away from $1$. There the signal term is
negligible against the noise floor, $z_{t,k} \approx (1-t)\,\varepsilon_k$, so the
noise is recoverable from the input, $\varepsilon_k \approx z_{t,k}/(1-t)$.
Substituting into \eqref{eq:vt},
\begin{equation}
v_{\text{target},k}
\;=\; x_k - \varepsilon_k
\;\approx\; -\varepsilon_k
\;\approx\; -\frac{z_{t,k}}{1-t}
\quad\Longrightarrow\quad
v_{\text{target}} \approx -\frac{z_t}{1-t}.
\label{eq:vt-heur}
\end{equation}
The target is then a deterministic function of the input: denoising is pure
rescaling and uses nothing about the data distribution.

\subsection{Algorithmic listings.}
\label{app:algorithm}
\cref{alg:sf_ct} defines the cutoff schedule $c(t)$ for each of the schedule shapes considered in \cref{sec:method:sf}. \cref{alg:sf_mask} constructs the soft 2D-DCT radial low-pass $M(t)$ from the scalar cutoff $c$. \cref{alg:sf_train} and \cref{alg:sf_sample} show how a training step and an Euler sampling step are modified by Spectral Forcing relative to the unmasked JiT recipe; class conditioning and CFG are omitted for brevity.

\begin{algorithm}[h]
\small
\caption{Spectral Forcing cutoff schedule $c(t)$.}
\label{alg:sf_ct}
\begin{Verbatim}[xleftmargin=1em,fontsize=\footnotesize]
# t in [0, 1]; c_min, c_max: cutoff bounds (defaults 0.05, 1.0)
# alpha: spectrum exponent (2.82 on ImageNet); eps = 1e-3 (endpoint guard)
# shape in {'linear', 'analytical', 'cosine', 't_squared', 't_sqrt'}
def c(t):
    if shape == 'linear':       f = t
    elif shape == 'cosine':     f = 0.5 - 0.5 * cos(pi * t)
    elif shape == 't_squared':  f = t ** 2
    elif shape == 't_sqrt':     f = sqrt(t)
    elif shape == 'analytical': 
        f = clip((eps / (1.0 - t).clamp_min(eps)) ** (2.0 / alpha), 0.0, 1.0)
    return c_min + (c_max - c_min) * f       # in [c_min, c_max]
\end{Verbatim}
\end{algorithm}

\begin{algorithm}[h]
\small
\caption{Soft 2D-DCT radial low-pass mask $M(t)$, given $c = c(t)$.}
\label{alg:sf_mask}
\begin{Verbatim}[xleftmargin=1em,fontsize=\footnotesize]
# c: scalar cutoff radius in [0, 1]
# H, W: image height, width (e.g. 256)
# kappa: soft transition sharpness (default 30)
def mask(c):
    u = arange(H); v = arange(W)              # DCT-II frequency indices
    U, V = meshgrid(u, v, indexing='ij')      # (H, W) integer grids
    r = sqrt(U**2 + V**2) / sqrt(2 * (W-1)**2) # normalized radius in [0, 1]
    return sigmoid(kappa * (c - r))            # soft low-pass at cutoff c
\end{Verbatim}
\end{algorithm}

\begin{algorithm}[h]
\small
\caption{Spectral Forcing training step.}
\label{alg:sf_train}
\begin{Verbatim}[xleftmargin=1em,fontsize=\footnotesize]
# net(z, t): diffusion transformer (e.g., JiT-700M/32)
# x: training batch
t = sample_t()
e = randn_like(x)
z = t * x + (1 - t) * e
v = (x - z) / (1 - t)
z_lp = idct(dct(z) * mask(c(t)))    # SF: input-side low-pass
x_pred = net(z_lp, t)
v_pred = (x_pred - z) / (1 - t)
loss = l2_loss(v - v_pred)
\end{Verbatim}
\end{algorithm}

\begin{algorithm}[h]
\small
\caption{Spectral Forcing sampling step.}
\label{alg:sf_sample}
\begin{Verbatim}[xleftmargin=1em,fontsize=\footnotesize]
# z: current samples at t
z_lp = idct(dct(z) * mask(c(t)))    # SF mask at current t
x_pred = net(z_lp, t)
v_pred = (x_pred - z) / (1 - t)
z_next = z + (t_next - t) * v_pred
\end{Verbatim}
\end{algorithm}

\begin{figure}[t]
\centering
\includegraphics[width=\textwidth]{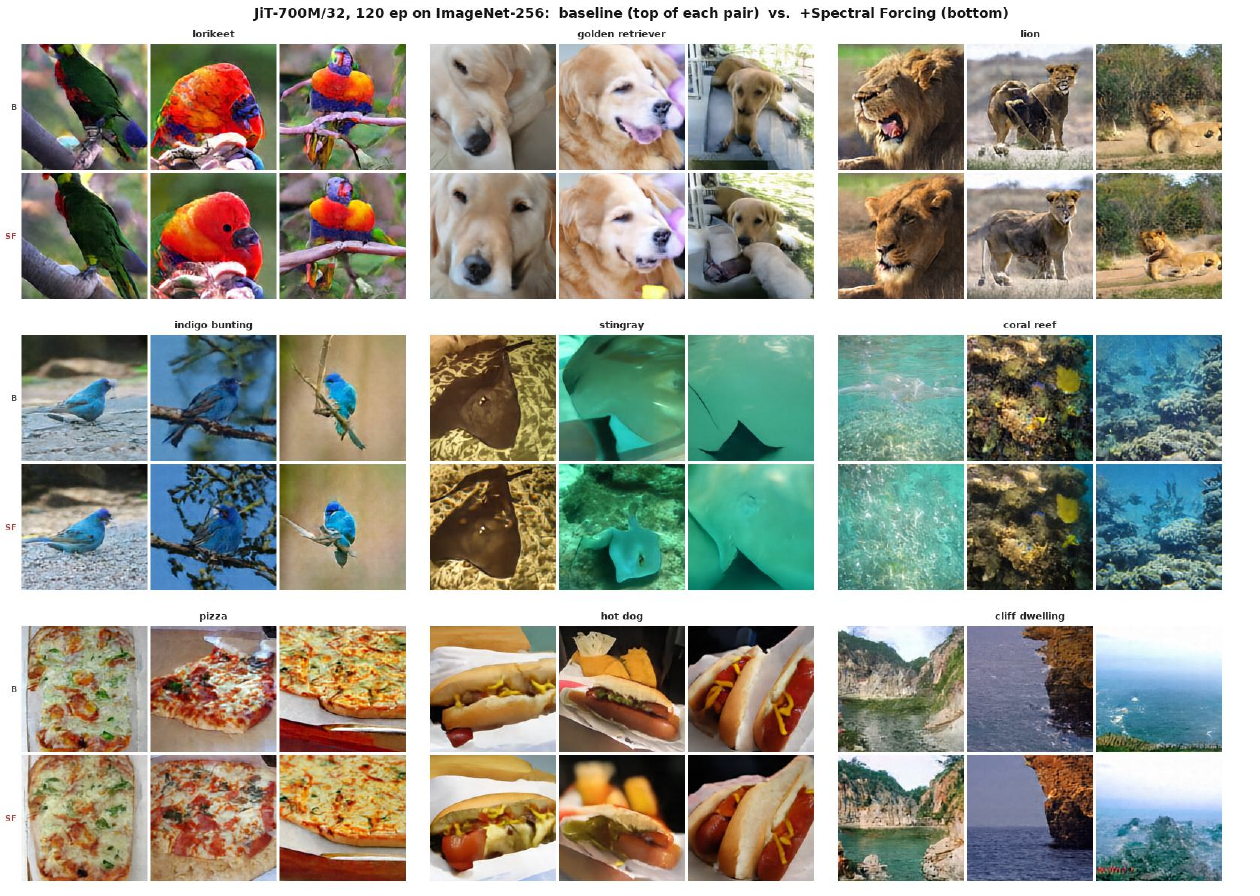}
\caption{\textbf{Qualitative samples on ImageNet-256.} \jit-700M/32 at $120$ epochs, baseline (\textbf{B}, top row of each block) vs.\ Linear-\sff{} (\textbf{SF}, bottom row), three sample indices per class, same class label and same sample index per column.}
\label{fig:qualitative_appendix}
\end{figure}

\subsection{Qualitative samples.}
\label{app:qualitative}
\cref{fig:qualitative_appendix} shows nine ImageNet classes generated by the same \jit-700M/32 model at $120$ epochs, comparing the no-mask baseline (FID $16.46$) against Linear-\sff{} (FID $\mathbf{15.15}$, $+8.0\%$). Each pair fixes both the class label and the sample index, so the only experimental variable is whether the time-conditional 2D-DCT low-pass was active during training and sampling. The classes span birds (lorikeet, indigo bunting), mammals (golden retriever, lion), marine subjects (stingray, coral reef), prepared food (pizza, hot dog), and a structured-scene class (cliff dwelling). Across all nine, \sff{} samples are visibly crisper in fine structure and more class-coherent at this converged budget; the effect is strongest on textured surfaces (lion mane, coral, pizza topping, cliff dwelling stonework) where the no-mask baseline tends to produce smoother but less specific texture.

\section{Limitations}
\label{app:limitations}

\paragraph{Per-step compute overhead.}
\sff{} adds one forward and one inverse 2D-DCT per denoising step --- approximately $0.5\%$ of per-step compute relative to the unmasked baseline at \jit-130M/32, $256^2$ (\cref{sec:discussion:efficiency}), with no learned parameters and no additional memory. The DCT is parallelizable on the patch grid and runs in the same kernel as patchify; for budgets where $0.5\%$ matters, the operator is a no-op to remove.

\paragraph{Hyperparameters held fixed across all ImageNet runs.}
The cutoff bounds $(c_{\min}, c_{\max}) {=} (0.05, 1.0)$ and the mask sharpness $\kappa{=}30$ are reused unchanged across every ImageNet configuration in \cref{tab:appendix_configs}, with no per-model, per-budget, or per-resolution tuning. The $c_{\min}$ sweep of \cref{app:cmin_sweep} indicates the operator is not narrowly tuned at this point; per-configuration tuning would only widen the reported gap to baseline.

\paragraph{Single benchmark and architecture family.}
We report on ImageNet-256, the canonical class-conditional FID/IS benchmark, and on \jit{}~\citep{li2025jit}, a modern pixel-space rectified-flow Transformer. \sff{} is parameter-free and recipe-agnostic by construction, so applying it to other backbones (e.g., DiT, U-Net) or other diffusion forwards is a configuration change, not an algorithmic change.


\end{document}